\documentclass{article}
\usepackage{enumitem}
\usepackage{microtype}
\usepackage{graphicx}
\usepackage{amsmath,amsfonts,amssymb,amsthm}
\usepackage[linesnumbered,ruled,vlined,boxed,algo2e]{algorithm2e}
\usepackage{mathtools}

\usepackage{commath}
\usepackage{subfigure}
\usepackage{multirow}
\usepackage{makecell}
\usepackage{booktabs} 
\usepackage{accents}
\usepackage{bm}
\usepackage{longtable}

\newcommand{\Te}{\emph{Teacher}}

\newcommand{\St}{\emph{Student}}

\usepackage{hyperref}
\usepackage[accepted]{conf}

\icmltitlerunning{Zero-Shot Knowledge Distillation in Deep  Networks}

\begin{document}

\twocolumn[
\icmltitle{Zero-Shot Knowledge Distillation in Deep  Networks}



\icmlsetsymbol{equal}{*}

\begin{icmlauthorlist}
\icmlauthor{Gaurav Kumar Nayak}{equal,iisc}
\icmlauthor{Konda Reddy Mopuri}{equal,edin}
\icmlauthor{Vaisakh Shaj}{equal,lincoln}
\icmlauthor{R. Venkatesh Babu}{iisc}
\icmlauthor{Anirban Chakraborty}{iisc}
\end{icmlauthorlist}

\icmlaffiliation{iisc}{Department of Computational and Data Sciences, Indian Institute of Science, Bangalore, India}
\icmlaffiliation{lincoln}{University of Lincoln, United Kingdom}
\icmlaffiliation{edin}{School of Informatics, University of Edinburgh, United Kingdom}

\icmlcorrespondingauthor{Gaurav Kumar Nayak}{gauravnayak@iisc.ac.in}

\icmlkeywords{Zero-shot, Knowledge Distillation, Pseudo Data, Data Impressions, Transfer Learning with Synthetic Data}

\vskip 0.3in
]



\printAffiliationsAndNotice{\icmlEqualContribution} 

\begin{abstract}
Knowledge distillation deals with the problem of training a smaller model (\St{}) from a high capacity source model (\Te{}) so as to retain most of its performance. Existing approaches use either the training data or meta-data extracted from it in order to train the \emph{Student}. However, accessing the dataset on which the \emph{Teacher} has been trained may not always be feasible if the dataset is very large or it poses privacy or safety concerns (e.g., bio-metric or medical data). Hence, in this paper, we propose a novel data-free method to train the \emph{Student} from the \textit{Teacher}. Without even using any meta-data, we synthesize the \textit{Data Impressions} from the complex \Te{} model and utilize these as surrogates for the original training data samples to transfer its learning to \St{} via knowledge distillation. We, therefore, dub our method ``Zero-Shot Knowledge Distillation" and demonstrate that our framework results in competitive generalization performance as achieved by distillation using the actual training data samples on multiple benchmark datasets.
\end{abstract}

\section{Introduction}
\label{sec:introduction}
Knowledge Distillation~\cite{hinton2015distilling} enables to transfer the complex mapping functions learned by cumbersome models to relatively simpler models. The cumbersome model can be an ensemble of multiple large models or a single model with large capacity and strong regualrizers such as Dropout~\cite{dropout-jmlr-2014}, BatchNorm~\cite{batchnorm-icml-2015}, etc. Typically the complex and small models are referred to as \emph{Teacher (T)} and \emph{Student (S)} models respectively. Generally the \Te{} models deliver excellent performance, but they can be huge and computationally expensive. Hence, these models can not be deployed in limited resource environments or when real-time inference is expected. On the other hand, a \St{} model has substantially less memory footprint, requires less computation, and thereby often results in a much faster inference time than that of the much larger \Te{} model. 

The latent information hidden in the confidences assigned by the \Te{} to the incorrect categories, referred to as `dark knowledge' is transferred to the \St{} via the distillation process. It is this knowledge that helps the \Te{} to generalize better and transfers to the \St{} via matching their soft-labels (output of the soft-max layer) instead of the one-hot vector encoded labels. Matching the soft-labels produced by the \Te{} is the natural way to transfer its generalization ability. For performing the knowledge distillation, one can use the training data from the target distribution or an arbitrary data. Typically, the data used to perform the distillation is called `Transfer set'. In order to maximize the information provided per sample, we can make the soft targets to have a high entropy (non-peaky). This is generally achieved by using a high temperature at the softmax layer~\cite{hinton2015distilling}. Also, because of non-peaky soft-labels, the training gradients computed on the loss will have less variance and enable to use higher learning rates leading to quick convergence.

The existing approaches use natural data either from the target data distribution or a different transfer set to perform the distillation. It is found by~\cite{hinton2015distilling} that using original training data performs relatively better. They also suggest to have an additional term in the objective for the \St{} to predict correct labels on the training data along with matching the soft-labels from the \Te{} (as shown in eq.~(\ref{eqn:kd})). However, accessing the samples over which the \Te{} had been trained may not always be feasible. Often the training datasets are too large (e.g., ImageNet~\cite{imagenet-ijcv-2015}). However, more importantly, most datasets are proprietary and not shared publicly due to privacy or confidentiality concerns.
Especially while dealing with biometric data of large population, healthcare data of patients etc. Also, quite often the corporate would not prefer its proprietary data to be potentially accessed by its competitors. In summary, data is more precious than anything else in the era of deep learning and hence access to premium data (used in training a model) may not always be realistic.

Therefore, in this paper, we present a novel data-free framework to perform knowledge distillation. Since we do not use any data samples (either from the target dataset or a different transfer set) to perform the knowledge transfer, we name our approach ``Zero-Shot Knowledge Distillation" (ZSKD). With no prior knowledge about the target data, we perform pseudo data synthesis from the \Te{} model that act as the transfer set to perform the distillation. Our approach obtains useful prior information about the underlying data distribution in the form of \emph{Class Similarities} from the model parameters of the \Te{}. Further, we successfully utilize this prior in the crafting process via modelling the output space of the \Te{} model as a Dirichlet distribution. We name the crafted samples \emph{Data Impressions} (DI) as these are the impressions of the training data as understood by the \Te{} model. Thus, the contributions of this work can be listed as follows:
\begin{itemize}
\item Unlike the existing methods that use either data samples or the extracted meta-data  to perform Knowledge Distillation, we present, for the first time, the idea of Zero-Shot Knowledge Distillation (ZSKD), with no data samples and no extracted prior information.
\item In order to compose a transfer set for performing distillation, we
present a sample extraction mechanism via modelling
the softmax space as a Dirichlet distribution and craft \emph{Data Impressions} (DI) from the parameters of a \Te{} model.
\item We present a simple, yet powerful procedure to extract useful prior in the form of \emph{Class Similarities} (sec.~\ref{subsec:dirichlet-modelling}) which enables better modelling of the data distribution and is utilized in the Dirichlet sampling based DI generation framework.
\item We demonstrate the effectiveness of our ZSKD approach via an empirical evaluation over multiple benchmark datasets and model architectures~(sec.~\ref{sec:experiments}).
\end{itemize}
The rest of the paper is organized as follows: section~\ref{sec:related-works} presents a brief account of existing research that are related to this work, section~\ref{sec:proposed-method} discusses the proposed framework in detail, section~\ref{sec:experiments} demonstrates the empirical evaluation, and section~\ref{sec:conclusion} presents a discussion on the proposed method and concludes the paper.
\section{Related Works}
\label{sec:related-works}
The teacher model generally has high complexity and are not preferred for real-time embedded platforms due to its large memory and computational requirements. In practice, networks of smaller size which are compact and deployable are required. Several techniques have been proposed in the past to transfer the knowledge from the teacher to the student model without much compromise in performance. We can categorize them broadly into three types based on the amount of data used for knowledge distillation:
\begin{itemize}
\item \textbf{Using entire training data or similar data}: In \cite{bucilua2006model}, model compression technique is used. The target network is trained using the pseudo labels obtained from the larger model with an objective to match the pre-softmax values (called logits). In \cite{hinton2015distilling}, the softmax distribution of classes produced by teacher model using high temperature in its softmax (called ``soft targets") are used to train the student model as the knowledge contained in incorrect class probabilities tends to capture the teacher generalization ability better in comparison to hard labels. The matching of logits is a special case of this general method. In \cite{furlanello2018born}, knowledge transfer is done across several generations where the student of current generation learns from its previous generation. The final predictions are made from the ensemble of student models using the mean of the predictions from each student. 
\item \textbf{Using few samples of original data}: In \cite{kimura2018few}, knowledge distillation is performed using few original samples of training data which are augmented by ``pseudo training examples". These pseudo examples are obtained using inducing point \cite{snelson2006sparse} method via iterative optimization technique in an adversarial manner which makes the training procedure complicated. 
\item \textbf{Using meta data}: In \cite{dfkd-nips-lld-17}, activation records are stored at each layer after the training of teacher model and used as meta data to reconstruct training samples and utilize them to train the student model. Although, this method does consider the case of knowledge distillation in the absence of training data but meta data is formed using the training data itself. So, meta data has dependency on training samples and hence it is not a complete data-free approach.
\end{itemize}
To the best of our knowledge, we are the first to demonstrate knowledge distillation in case where no training data is available in any form. It has been shown by \cite{mopuri2018ask} that the pretrained models have memory in terms of learned parameters and can be used to extract class representative samples. Although, it was used in the context of adversarial perturbation task, we argue that carefully synthesized samples can be used as pseudo training data for knowledge distillation.
\section{Proposed Method}
\label{sec:proposed-method}
In this section, we briefly introduce the process of Knowledge Distillation (KD) and present the proposed framework for performing Zero-Shot Knowledge Distillation in detail.
\subsection{Knowledge Distillation}
\label{subsec:kd}
Transferring the generalization ability of a large, complex \Te{} $(T)$ deep neural network to a less complex \St{} $(S)$ network can be achieved using the class probabilities produced by a \Te{} as ``soft targets"~\cite{hinton2015distilling} for training the \St{}. For this transfer, existing approaches require access to the original training data consisting of tuples of input data and targets $(x,y) \in \mathbb{D}$. Let $T$ be the \Te{} network with learned parameters $\theta_T $ and $S$ be the \St{} with parameters $\theta_S$, note that in general $\lvert \theta_S \rvert \ll \lvert \theta_T \rvert$. Knowledge distillation methods train the \St{} via minimizing the following objective $(L)$ with respect to the parameters $\theta_S$ over the training samples ${(x,y) \in \mathbb{D}}$
\begin{equation}
    L = \sum_{ (x,y) \in \mathbb{D} }L_{KD}(S(x,\theta_S,\tau),T(x,\theta_T,\tau))  +  \lambda  L_{CE}({\hat{y}}_S, y)
    \label{eqn:kd}
\end{equation}
 $L_{CE}$ is the cross-entropy loss computed on the labels ${\hat{y}}_S$ predicted by the \St{} and their corresponding ground truth labels $y$. $L_{KD}$ is the distillation loss (e.g. cross-entropy or mean square error) comparing the soft labels (softmax outputs) predicted by the \St{} against the soft labels predicted by the \Te{}. $T(x,\theta_T)$ represents the softmax output of the \Te{} and $S(x,\theta_S)$ denotes the softmax output of the \St{}. Note that, unless it is mentioned, we use a softmax temperature of $1$. If we use a temperature value $(\tau)$ different from $1$, we represent it as $S(x,\theta_S,\tau)$ and $T(x,\theta_T,\tau)$ for the remainder of the paper. $\lambda$ is the hyper-parameter to balance the two objectives.
\subsection{Modelling the Data in Softmax Space} 
\label{subsec:dirichlet-modelling}
However, in this work, we deal with the scenario where we have no access to (i) any training data samples (either from the target distribution or different), or (ii) meta-data extracted from it~(e.g. \cite{dfkd-nips-lld-17}). In order to tackle this, our approach taps the memory (learned parameters) of the \Te{} and synthesizes pseudo samples from the underlying data distribution on which it is trained. Since these are the impressions of the training data extracted from the trained model, we name these synthesized input representations as \emph{Data Impressions}. We argue that these can serve as representative samples from the training data distribution, which can then be used as a transfer set in order to perform the knowledge distillation to a desired \St{} model.

Thus, in order to craft the \emph{Data Impressions}, we model the output (softmax) space of the \Te{} model. Let $\bm{s}\sim p(\bm{s})$, be the random vector that represents the neural softmax outputs of the \Te{}, $T(x,\theta_T)$. We model $p(\bm{s}^k)$ belonging to each class $k$, using a Dirichlet distribution which is a distribution over vectors whose components are in $[0,1]$ range and their sum is $1$. Thus, the distribution to represent the softmax outputs  $\bm{s}^k$ of class $k$ would be modelled as, $Dir(K, \bm{\alpha}^k )$, where $k \in \{1 \ldots K\}$ is the class index, $K$ is the dimension of the output probability vector (number of categories in the recognition problem) and $\bm{\alpha}^k$ is the concentration parameter of the distribution modelling class $k$. The concentration parameter $\bm{\alpha}^k$ is a $K$ dimensional positive real vector, i.e, $\bm{\alpha}^k$ $= [\alpha^k_1,\alpha^k_2,\ldots,\alpha^k_K ], \text{and} ~ \alpha^k_i >0, \forall i$.

\textbf{Concentration Parameter ($\bm{\alpha}$)}: Since the sample space of the Dirichlet distribution is interpreted as a discrete probability distribution (over the labels), intuitively, the concentration parameter $(\bm{\alpha})$ can be thought of as determining how ``concentrated" the probability mass of a sample from a Dirichlet distribution is likely to be. With a value much less than $1$, the mass will be highly concentrated in only a few components, and all the rest will have almost zero mass. On the other hand, with a value much greater than $1$, the mass will be dispersed almost equally among all the components.

Obtaining prior information for the concentration parameter is not straightforward. The parameter cannot be the same for all components since this results in all sets of probabilities being equally likely, which is not a realistic scenario. For instance, in case of CIFAR-$10$ dataset, it would not be meaningful to have a softmax output in which the \emph{dog} class and \emph{plane} class have the same confidence (since they are visually dissimilar). Also, same $\alpha_i$ values denote the lack of any prior information to favour one component of sampled softmax vector over the other. Hence, the concentration parameters should be assigned in order to reflect the similarities across the components in the softmax vector. Since these components denote the underlying categories in the recognition problem, $\bm{\alpha}$ should reflect the \emph{visual} similarities among them.

Thus, we resort to the \Te{} network for extracting this information. We compute a normalized class similarity matrix $(C)$ using the weights $W$ connecting the final (softmax) and the pre-final layers. The element $C(i,j)$ of this matrix denotes the visual similarity between the categories $i$ and $j$ in $[0,1]$. Thus, a row $\bm{c}_k$ of the class similarity matrix $(C)$ gives the similarity of class $k$ with each of the $K$ categories (including itself). Each row $\bm{c}_k$ can be treated as the concentration parameter $({\bm\alpha})$ of the Dirichlet distribution $(Dir)$, which models the distribution of output probability vectors belonging to class $k$.

\begin{figure}[!]
  \centering
  \includegraphics[width=0.5\textwidth]{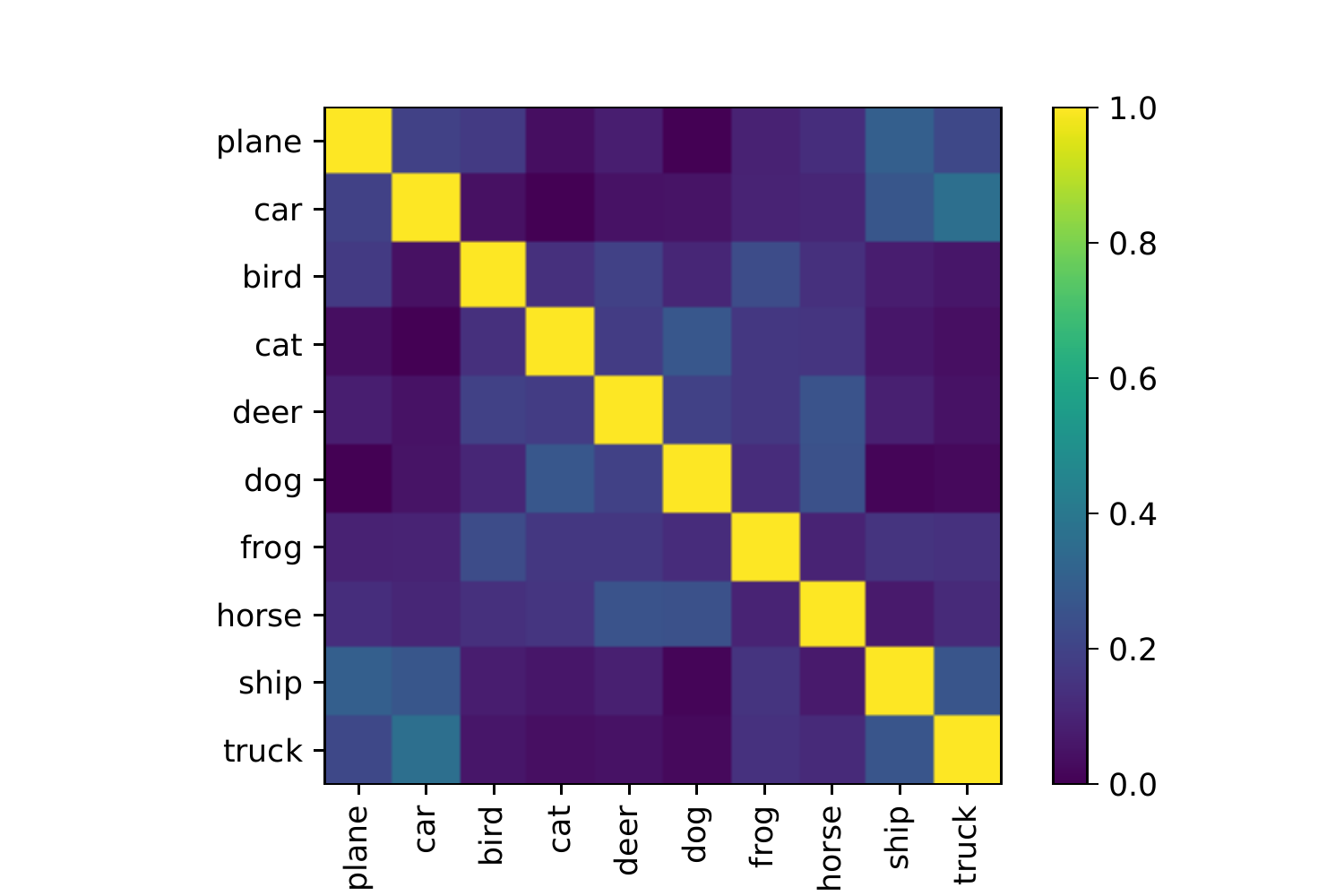}%
  \caption{Class similarity matrix computed for the \Te{} model trained over CIFAR-$10$ dataset. Note that the class labels are mentioned and the learned similarities are meaningful.}
  \label{fig:similarity-matrix-cifar}
  \end{figure}

\textbf{Class Similarity Matrix}: The class similarity matrix $C$ is calculated as follows. The final layer of a typical recognition model will be a fully connected layer with a softmax non-linearity. Each neuron in this layer corresponds to a class $(k)$ and its activation is treated as the probability predicted by the model for that class.  The weights connecting the previous layer to this neuron $(\bm{w}_k)$ can be considered as the template of the class $k$ learned by the \Te{} network. This is because the predicted class probability is proportional to the alignment of the pre-final layer's output with the template $(\bm{w}_k)$. The predicted probability peaks when the pre-final layer's output is a positive scaled version of this template $(\bm{w}_k)$. On the other hand, if the output of the pre-final layer is misaligned with the template $\bm{w}_k$, the confidence predicted for class $k$ is reduced. Therefore, we treat the weights $\bm{w}_k$ as the class template for class $k$ and compute the similarity between classes $i$ and $j$ as:
\begin{equation}
    C(i,j)=\frac{\bm{w}_i^T\bm{w}_j}{ \norm{\bm{w}_i} \norm{\bm{w}_j} }
    \label{eqn:class-sim-mat}
\end{equation}
Since the elements of the concentration parameter  have to be positive real numbers, we further perform a min-max normalization over each row of the class similarity matrix. The visualization of the class similarity matrix calculated from a CIFAR-$10$ trained model is shown in Figure~\ref{fig:similarity-matrix-cifar}.
\subsection{Crafting \textit{Data Impressions} via Dirichlet Sampling}
\label{subsec:crafting}
Once the parameters $K$ and $\bm{\alpha}^k$ of the Dirichlet distribution are obtained for each class $k$, we can sample class probability (softmax) vectors, which respect the class similarities as learned by the \Te{} network. Using the optimization procedure in eq.~(\ref{eqn:dir-ci}) we obtain the input representations corresponding to these sampled output class probabilities.
Let $Y^{k} = [\bm{y}_1^{k},\bm{y}_2^{k},\ldots,\bm{y}_N^{k}] \in \ \mathbb{R}^{K \times N}$, be the $N$ softmax vectors corresponding to class $k$, sampled from $Dir(K,\bm{\alpha}^k)$ distribution. Corresponding to each sampled softmax vector $\bm{y}_i^{k}$, we can craft a \emph{Data Impression} ${\bar{x}_i}^k$, for which the \Te{} predicts a similar softmax output. We achieve this by optimizing the objective shown in eq.~(\ref{eqn:dir-ci}). We initialize $\bar{x}^k_i$ as a random noisy image and update it over multiple iterations till the cross-entropy loss between the sampled softmax vector $(\bm{y}_i^{k})$ and the softmax output predicted by the \Te{} is minimized.
\begin{equation}
    \bar{x_i}^k=\underset{x}{\mathrm{argmin}} \ L_{CE}(\bm{y}_i^{k},T(x,\theta_T,\tau))
    \label{eqn:dir-ci}
\end{equation}
where $\tau$ is the temperature used in the softmax layer. The process is repeated for each of the $N$ sampled softmax probability vectors in $Y^{k}$, $k \in \{1 \ldots K\}$.

\textbf{Scaling Factor ($\beta$)}: The probability density function of the Dirichlet distribution for $K$ random variables is a $K-1$ dimensional probability simplex that exists on a $K$ dimensional space. In addition to parameters $K$ and $\bm{\alpha}$ as discussed in section~\ref{subsec:dirichlet-modelling}, it is important to discuss the significance of the range of $\alpha_i  \in  \bm{\alpha}$ , in controlling the density of the distribution. When $\alpha_i < 1, \  \forall  i \ \epsilon[1,K]$, the density congregates at the edges of the simplex~\cite{balakrishnan2004primer,lin2016dirichlet}. As their values increase (when $\alpha_i > 1,  \forall i \in [1,K]$), the density becomes more concentrated on the center of the simplex~\cite{balakrishnan2004primer,lin2016dirichlet}. Thus, we define a scaling factor $(\bm{\beta})$ which can control the range of the individual elements of the concentration parameter, which in turn decides regions in the simplex from which sampling is performed. This becomes a hyper-parameter for the algorithm. Thus the actual sampling of the probability vectors happen from $p(\bm{s})= Dir( K, \beta \times \bm{\alpha})$. $\beta$ intuitively models the spread of the Dirichlet distribution and acts as a scaling parameter atop $\bm\alpha$ to yield the final concentration parameter (prior). $\beta$ controls the $l_1$-norm of the final concentration parameter which, in turn, is inversely related to the variance of the distribution. Variance of the sampled simplexes is high for smaller values of $\beta$ . However very low values for $\beta$  (e.g. $0.01$), in conjunction with the chosen $\bm\alpha$, result in highly sparse softmax vectors concentrated on the extreme corners of the simplex, which is equivalent to generating class impressions (see Fig.~\ref{fig:ci-vs-di}). As per the ablation studies, $\beta$ values of 0.1, 1.0 or a mix of these are in general favorable since they encourage higher diversity (variance) and at the same time does not result in highly sparse vectors. 
\subsection{Zero-Shot Knowledge Distillation}
\label{subsec:zskd}
\begin{figure*}[!]
\centering
 \includegraphics[width=\textwidth, height=0.4\textwidth]{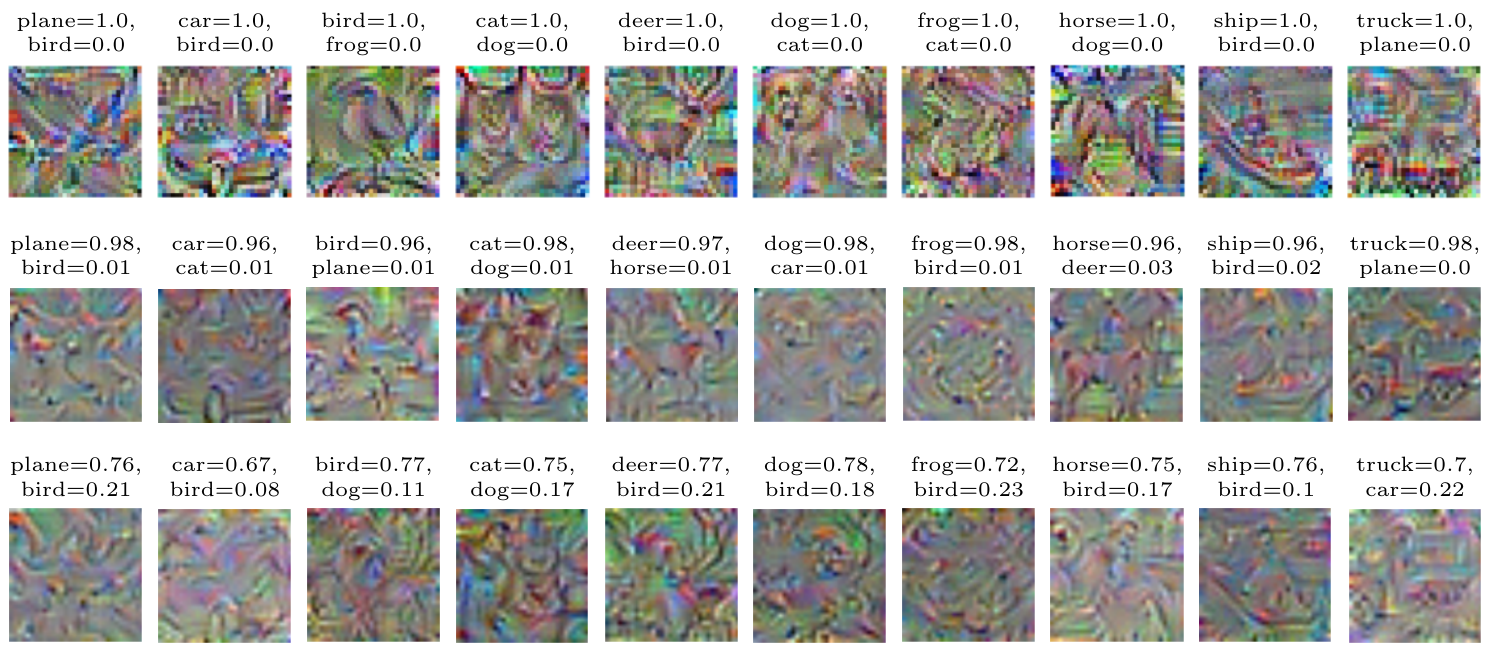} \hfill
\caption{Visualizing the \emph{DI}s synthesized from the \Te{} model trained on the CIFAR-$10$ dataset for different choices of output softmax vectors (i.e., output class probabilities). Note that the figure shows $3$ \emph{DI}s per class in each column, each having a different spread over the labels. However, only the top-$2$ confidences in the sampled softmax corresponding to each $DI$ are mentioned on top for clarity. Please note that there is no explicit objective for encouraging these pseudo samples to be visually closer to the actual training data samples, and yet some of the samples show striking patterns visually very similar to actual object shapes (e.g., bird, car, cat, dog, and deer in the first/second rows).}
\label{fig:vis-mnist}
\end{figure*}

Once we craft the Data Impressions (DI) $(\bar{X})$ from the \Te{} model, we treat them as the `Transfer set' and perform the knowledge distillation. Note that we use only the distillation loss $L_{KD}$ as shown in eq.~(\ref{eqn:zskd}). We ignore the cross-entropy loss from the general Distillation objective (eq.~(\ref{eqn:kd})) since there is only minor to no improvement in the performance and it reduces the burden of hyper-parameter $\lambda$. The proposed ZSKD approach is detailed in Algorithm~\ref{algo:zskd}.
\begin{align}
    \theta_S=\underset{\theta_S}{\mathrm{\:argmin\:}} \sum_{\bar{x} \in \bar{X}} L_{KD}(T(\bar{x},\theta_T,\tau), S(\bar{x},\theta_S,\tau))
    \label{eqn:zskd}
\end{align}
Thus we generate a diverse set of pseudo training examples that can provide with enough information to train the \St{} model via Dirichlet sampling. Some of the \emph{Data Impressions} are presented in Figure~\ref{fig:vis-mnist} 
for CIFAR-$10$ dataset. Note that the figures show $3$ \emph{DI}s per category. Also, note that the top-$2$ confidences in the sampled softmax corresponding to each $DI$ are mentioned on top. We observe that the \emph{DI}s are visually far away from the actual data samples of the dataset. However, some of the \emph{DI}s synthesized from peaky softmax vectors (e.g. the bird, cat, car, and deer in the first row) contain clearly visible patterns of the corresponding objects. The observation that the \emph{DI}s being visually far away from the actual data samples is understandable, since the objective to synthesize them (eq.~(\ref{eqn:dir-ci})) pays no explicit attention to visual detail. 
\begin{algorithm}[t]
\SetAlgoLined
\SetKwInOut{Input}{Input}  
\Input{\Te{} model $T$ \\$N$: number of DIs crafted per category,\\ $[\beta_1,\beta_2,...,\beta_B]$: $B$ scaling factors,\\
$\tau$: Temperature for distillation 
}
\SetKwInOut{Output}{Output}  
\Output{Learned \St{} model $S(\theta_S)$,\\ $\bar{X}$: \emph{Data Impressions}}
Obtain $K$: number of categories from \emph{T}

Compute the class similarity matrix\\ $C=[\textbf{c}_1^T,\textbf{c}_2^T,\ldots,\textbf{c}_K^T]$ as in eq.~(\ref{eqn:class-sim-mat})

$\bar{X} \leftarrow \emptyset$

\For{k=1:K }{

Set the concentration parameter $\bm{\alpha}^k =\textbf{c}_k$

    \For{b=1:B}{

        \For{n=1:$\left \lfloor{N/B}\right \rfloor $}{
        
            Sample ${\bm{y}_n^k} \sim Dir(K,\beta_b \times \bm{\alpha}^k)$

            Initialize $\bar{x}_n^k$ to random noise and craft $\bar{x}_n^k = \underset{x}{\mathrm{argmin}} \ L_{CE}({\bm{y}_n^k},T(x,\theta_T,\tau))$

            $\bar{X} \gets \bar{X} \cup \bar{x}_n^k$ 
            
        }

    }
    
}
Transfer the \Te{}'s knowledge to \St{} using the \emph{DIs} via  $\theta_S=\underset{\theta_S}{\mathrm{\:argmin\:}} \sum_{\bar{x} \in \bar{X}} L_{KD}(T(\bar{x},\theta_T,\tau), S(\bar{x},\theta_S,\tau))$ 
\caption{Zero-Shot Knowledge Distillation}
\label{algo:zskd}
\end{algorithm}
\section{Experiments}
\label{sec:experiments}
In this section, we discuss the experimental evaluation of the proposed data-free knowledge transfer framework over a set of benchmark object recognition datasets: MNIST \cite{lecun1998gradient}, Fashion MNIST (FMNIST) \cite{xiao2017fashion}, and CIFAR-$10$ \cite{krizhevsky2009learning}. As all the experiments in these three datasets are dealing with classification problems with $10$ categories each, value of the parameter $K$ in all our experiments is $10$. For each dataset, we first train the \Te{} model over the available training data using the cross-entropy loss. Then we extract a set of \emph{Data Impressions} $(DI)$ from it via modelling its softmax output space as explained in sections~\ref{subsec:dirichlet-modelling} and \ref{subsec:crafting}. Finally, we choose a (light weight) \St{} model and train over the transfer set (DI) using eq.~(\ref{eqn:zskd}). 

We consider two $(B=2)$ scaling factors, $\beta_{1} = 1.0$ and $\beta_{2} = 0.1$ across all the datasets, i.e., for each dataset, half the \emph{Data Impressions} are generated with $\beta_1$ and the other with $\beta_2$. However we observed that one can get a fairly decent performance with a choice of beta equal to either 0.1 or 1 (even without using the mixture of Dirichlet) across the datasets. A temperature value $(\tau)$ of $20$ is used across all the datasets. We investigate (in sec.~\ref{subsec:effect-size}) the effect of transfer set size, i.e., the number of \emph{Data Impressions} on the performance of the \St{} model. Also, since the proposed approach aims to achieve better generalization, it is a natural choice to augment the crafted \emph{Data Impressions} while performing the distillation. We augment the samples using regular operations such as scaling, translation, rotation, flipping etc. which has proven useful in further boosting the model performance \cite{dao2018kernel}. Please note that the appendix of the paper provides various information such as the exact architectural details, hyper-parameters used, and the augmentations performed on the \emph{Data Impressions}. The codes of the project are available at \url{https://github.com/vcl-iisc/ZSKD}.
\begin{table}[]
\caption{Performance of the proposed ZSKD framework on the MNIST dataset.}
\label{tab:mnist}
\centering
\begin{tabular}{|c|c|}
\hline
\multicolumn{1}{|c|}{\textbf{Model}} & \multicolumn{1}{c|}{\textbf{Performance}} \\ \hline
Teacher-CE                             &       99.34                                    \\ \hline
Student-CE                           &       98.92                                      \\ \hline
\makecell{Student-KD \cite{hinton2015distilling}\\ 60K original data}           &       99.25                                    \\ \hline
\makecell{\cite{kimura2018few} \\200 original data}                           &   86.70                                        \\ \hline
\makecell{\cite{dfkd-nips-lld-17}\\ (uses meta data)}                           &    92.47                                       \\ \hline
\makecell{\textbf{ZSKD} (Ours) \\($24000$ \emph{DI}s, and no original data)}                         &       98.77                                    \\ \hline
\end{tabular}
\end{table}
\subsection{MNIST}
\label{subsec:mnist}
The MNIST dataset has $60000$ training images and $10000$ test images of handwritten digits. We consider Lenet-$5$ for the \Te{} model and Lenet-$5$-Half for \St{} model similar to \cite{dfkd-nips-lld-17}. The Lenet-$5$ Model contains $2$ convolution layers and pooling which is followed by three fully connected layers.  Lenet-$5$ is modified to make Lenet-$5$-Half by taking half the number of filters in each of the convolutional layers. The \Te{} and \St{} models have $61706$ and $35820$ parameters respectively. Input images are resized from $28 \times 28$ to $32 \times 32$ and the pixel values are normalized to be in $[0, 1]$ before feeding into the models. 

The performance of our Zero-Shot Knowledge Distillation for MNIST dataset is presented in Table~\ref{tab:mnist}. Note that, in order to understand the effectiveness of the proposed ZSKD, the table also shows the performance of the \Te{} and \St{} models trained over actual data samples along with a comparison against existing distillation approaches. Teacher-CE denotes the classification accuracy of the \Te{} model trained using the cross-entropy (CE) loss, Student-CE denotes the performance of the \St{} model trained with all the training samples and their ground truth labels using cross-entropy loss. Student-KD denotes the accuracy of the \St{} model trained using the actual training samples through Knowledge Distillation (KD) from \Te{}. Note that this result can act as a vague upper bound for the data-free distillation approaches.

It is clear that the proposed Zero-Shot Knowledge Distillation (ZSKD) outperforms the existing few data \cite{kimura2018few} and data-free counterparts \cite{dfkd-nips-lld-17} by a great margin. Also, it performs close to the full data (classical) Knowledge Distillation while using only $24000$ \textit{DI}s, i.e., $40\%$ of the the original training set size.
\subsection{Fashion MNIST}
\label{subsec:fmnist}
In comparison to MNIST, this dataset is more challenging and contains images of fashion products. The training and testing set has $60000$ and $10000$ images respectively. Similar to MNIST, we consider Lenet-$5$ and Lenet-$5$-Half as \Te{} and \St{} model respectively where each input image is resized from dimension $28 \times 28$ to $32 \times 32$. 

Table~\ref{tab:fmnist} presents our results and compares with the existing approaches. Similar to MNIST, ZSKD outperforms the existing few data knowledge distillation approach~\cite{kimura2018few} by a large margin, and performs close to the classical knowledge distillation scenario~\cite{hinton2015distilling} with all the training samples.
\begin{table}[]
\caption{Performance of the proposed ZSKD framework on the Fashion MNIST dataset.}
\label{tab:fmnist}
\centering
\begin{tabular}{|c|c|}
\hline
\multicolumn{1}{|c|}{\textbf{Model}} & \multicolumn{1}{c|}{\textbf{Performance}} \\ \hline
Teacher-CE                              &     90.84                                      \\ \hline
Student-CE                           &     89.43                                     \\ \hline
\makecell{Student-KD \cite{hinton2015distilling}\\ 60K original data}          &     89.66                                      \\ \hline
\makecell{\cite{kimura2018few} \\200 original data}                        &                 72.50                           \\ \hline
\makecell{\textbf{ZSKD} (Ours) \\($48000$ \emph{DI}s, and no original data)}                          &     79.62                                     \\ \hline
\end{tabular}
\end{table}

\begin{figure*}[h]
\centering
\noindent\begin{minipage}{\textwidth}
  \centering
  \includegraphics[width=.31\textwidth]{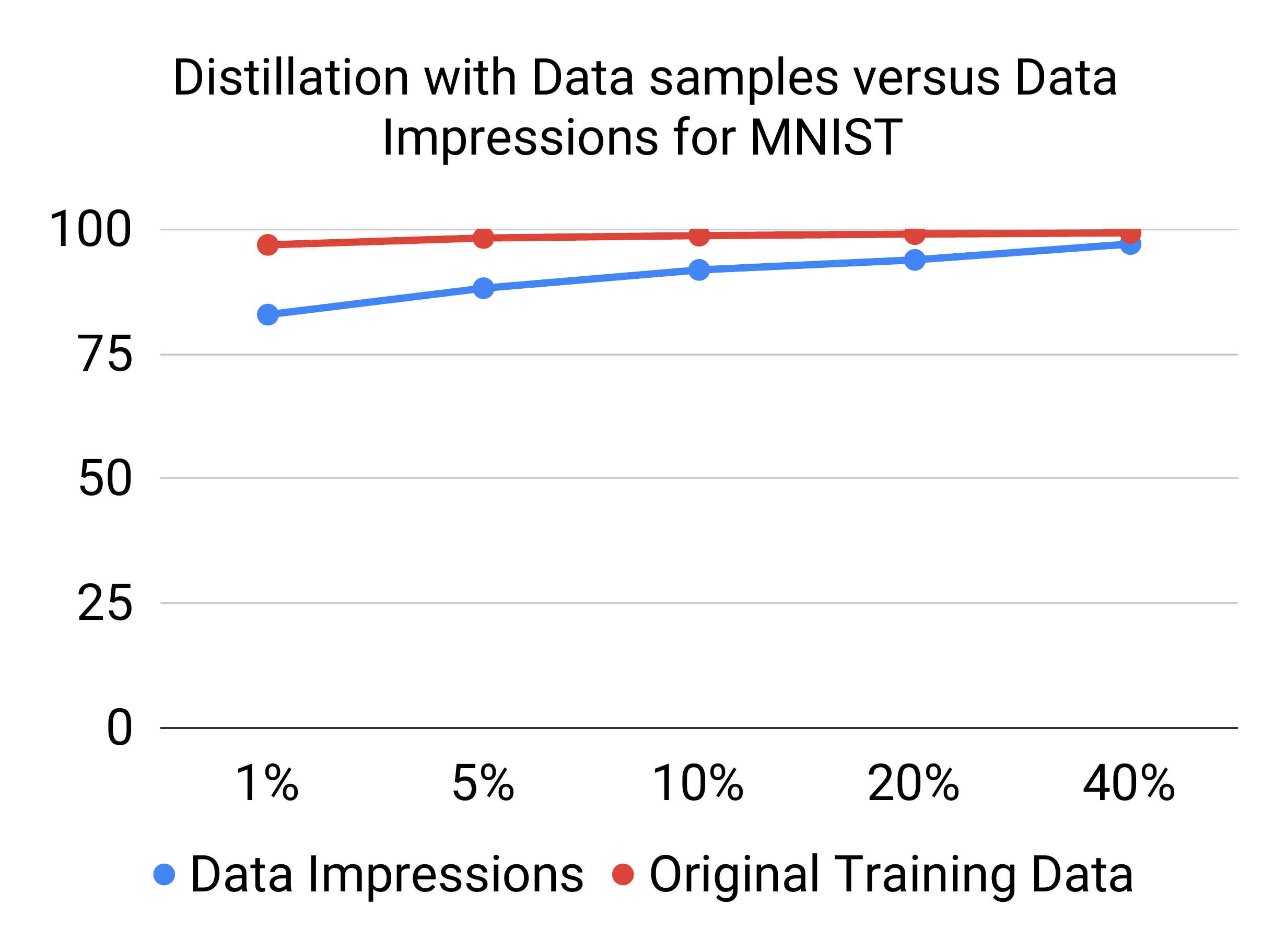}\hfill
  \includegraphics[width=.31\textwidth]{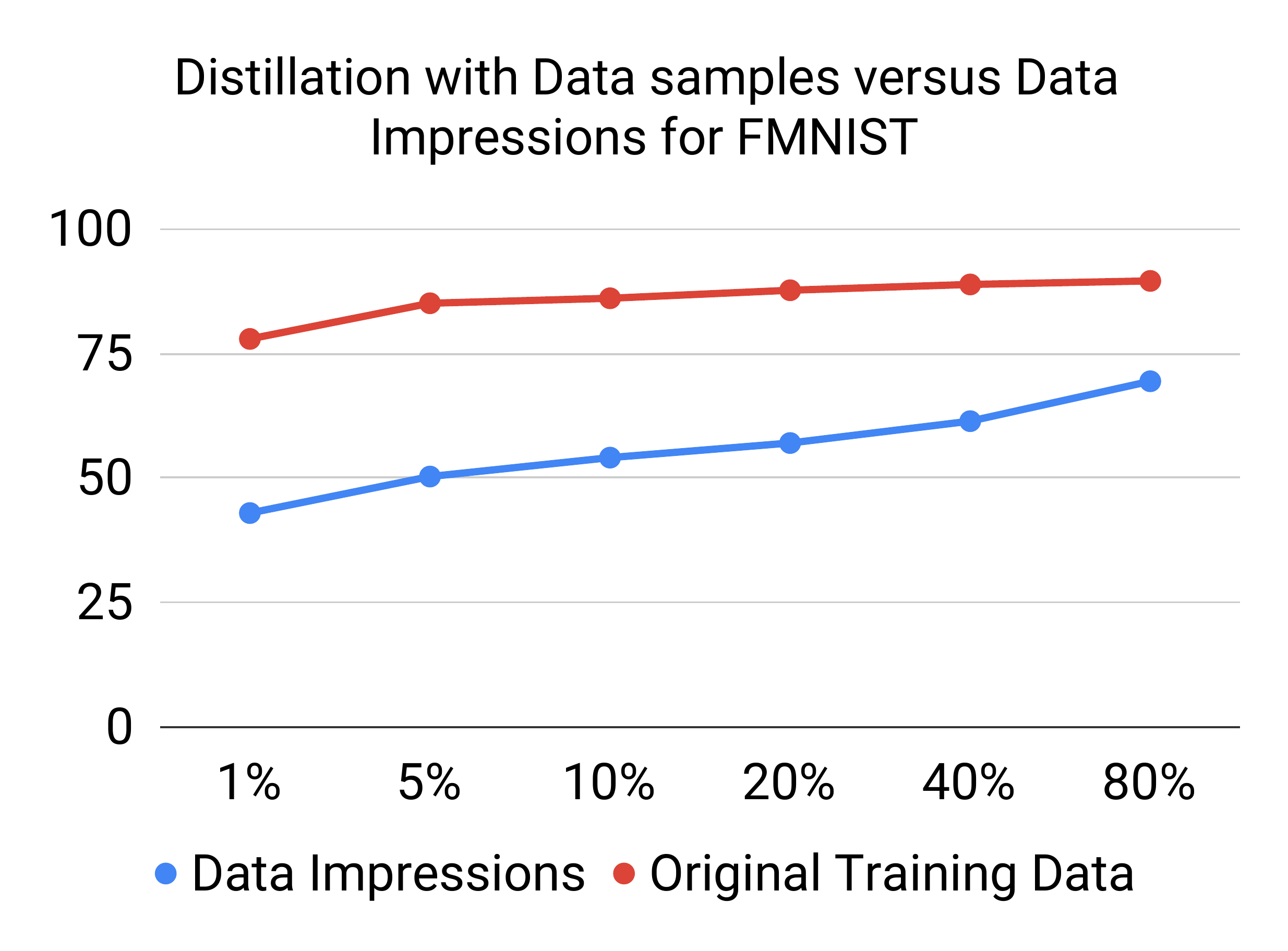}\hfill
  \includegraphics[width=.31\textwidth]{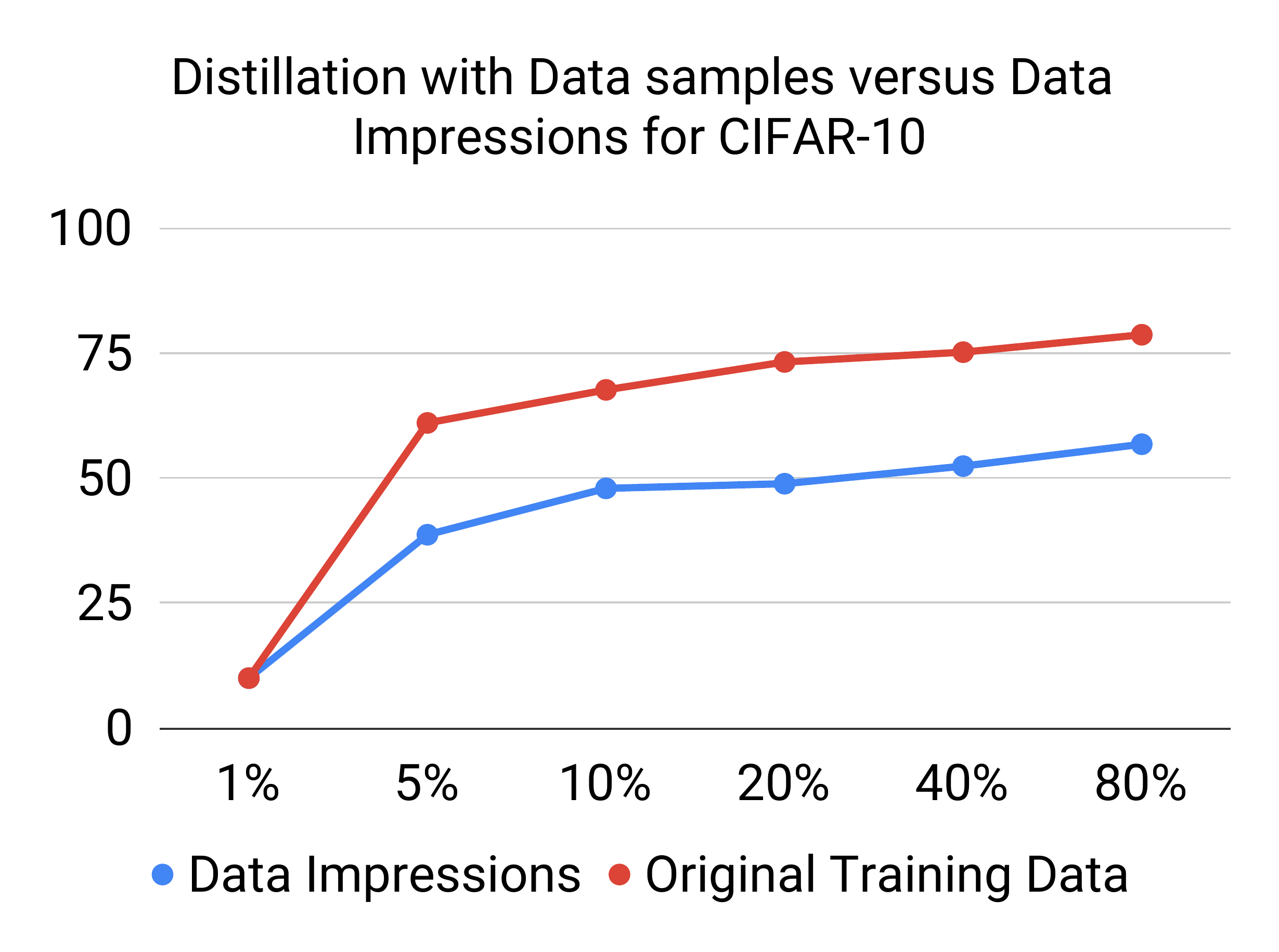}\hfill
  \vspace{0.01\textwidth}
\end{minipage}
\caption{Performance (Test Accuracy) comparison of Data samples versus Data Impressions (without augmentation). Note that the x-axis denotes the number of \emph{DI}s or original training samples (in $\%$) used for performing Knowledge Distillation with respect to the size of the training data.
}
\label{fig:data-vs-di}
\end{figure*}
\begin{figure*}[]
\centering
\noindent\begin{minipage}{\textwidth}
  \centering
  \includegraphics[width=.32\textwidth]{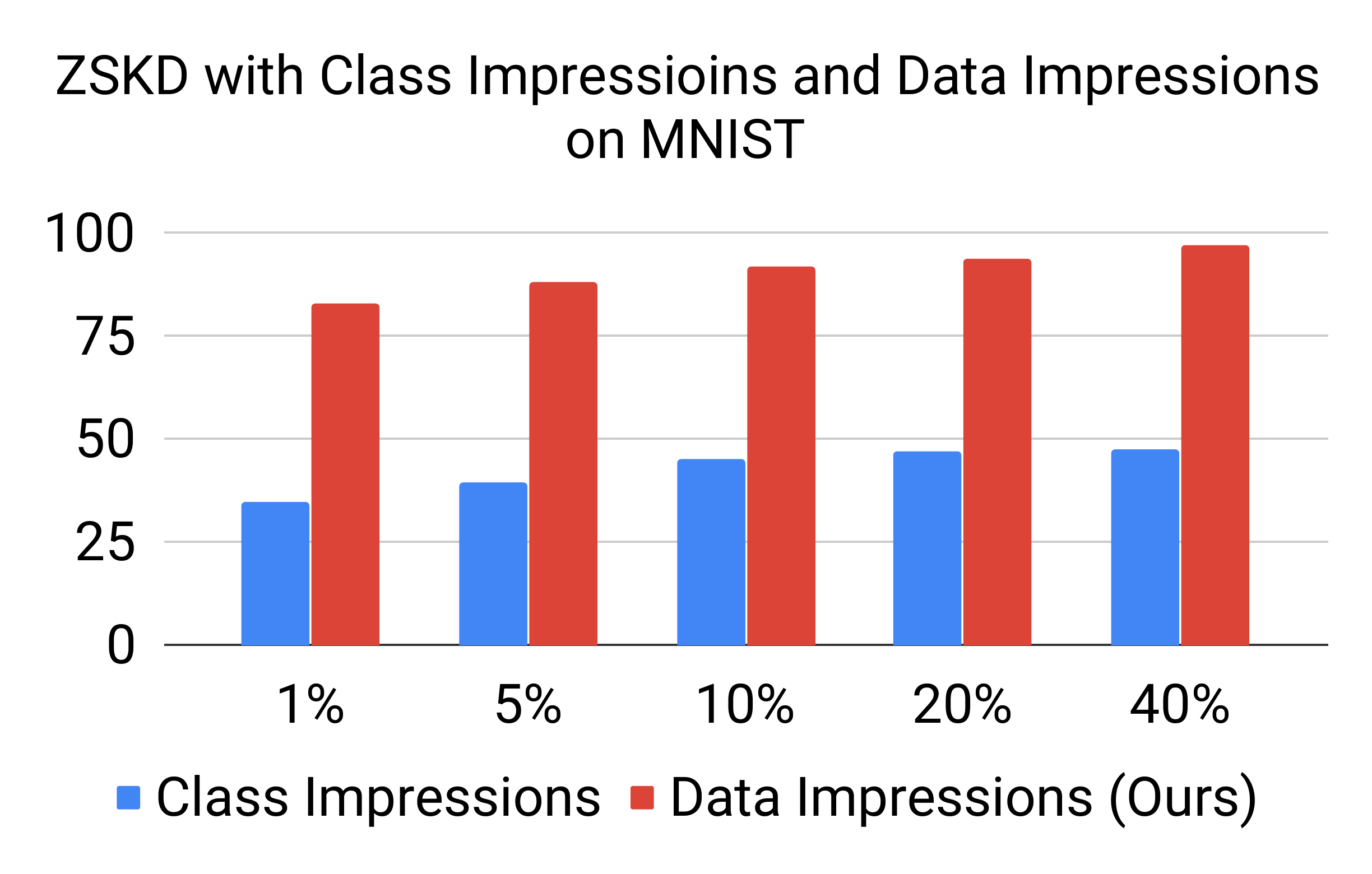}\hfill
  \includegraphics[width=.32\textwidth]{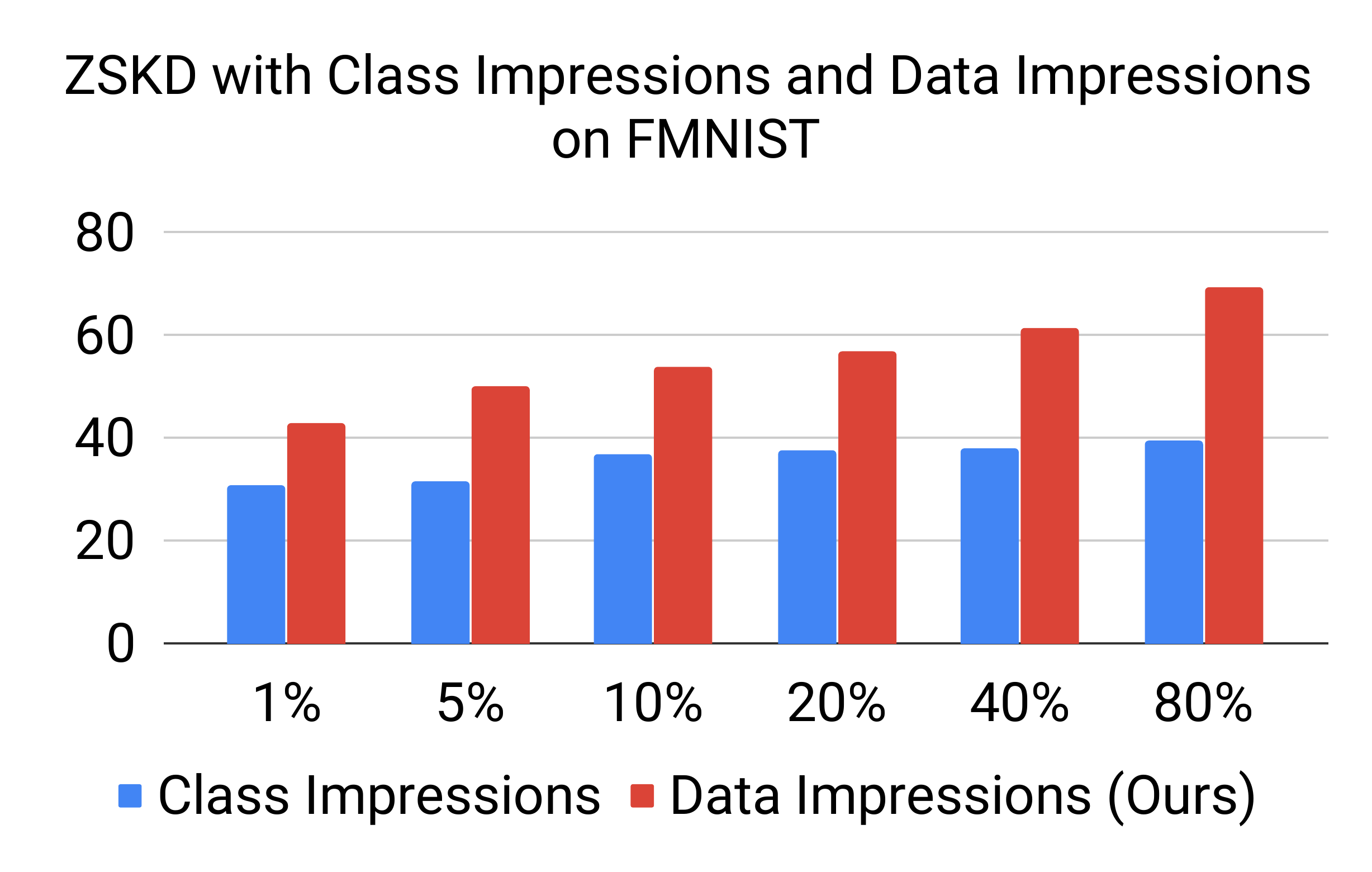}\hfill
  \includegraphics[width=.32\textwidth]{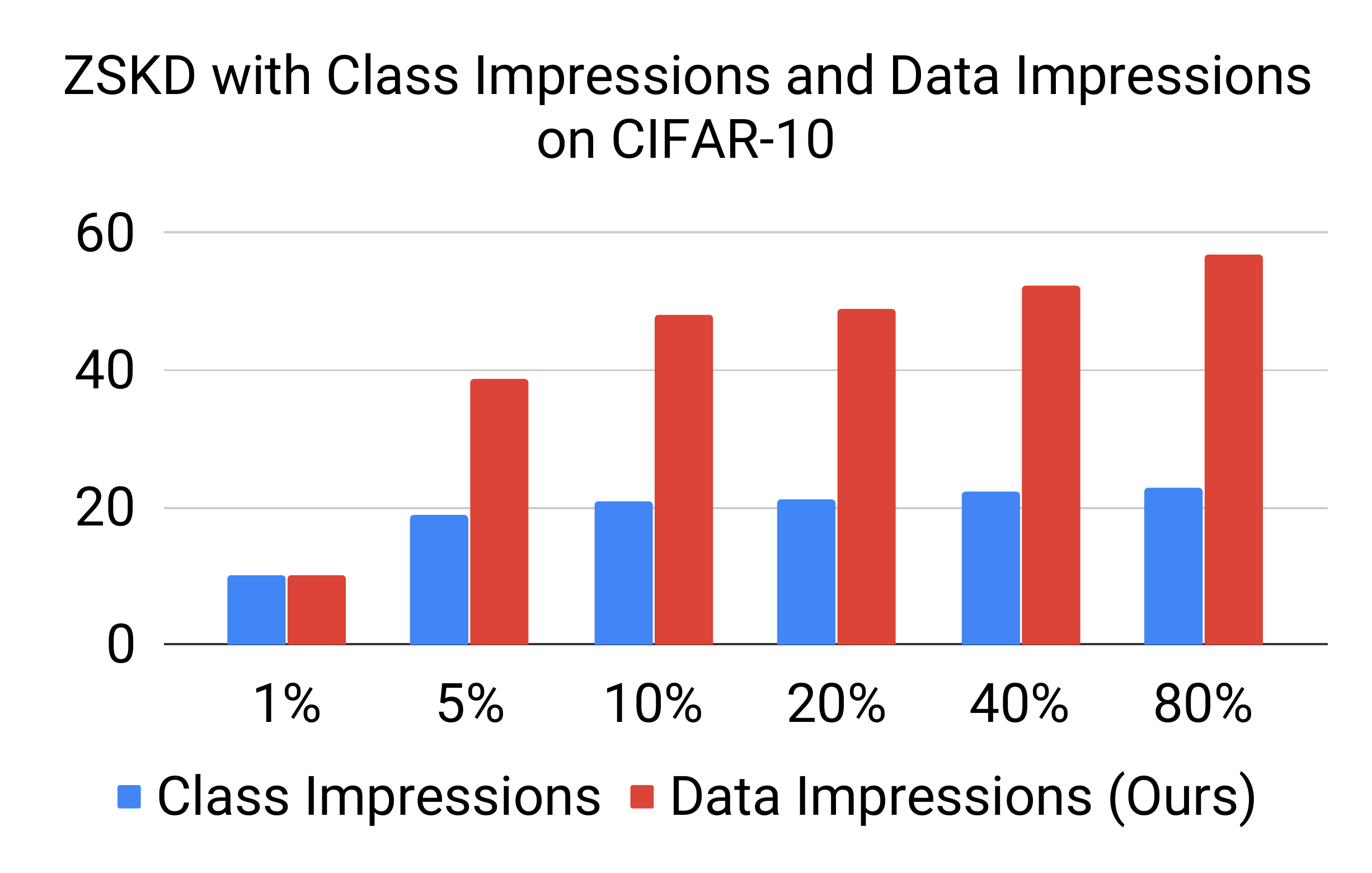}\hfill
  \vspace{0.01\textwidth}
\end{minipage}
\caption{Performance (Test Accuracy) comparison of the ZSKD with Class Impressions~\cite{mopuri2018ask} and proposed Data Impressions (without augmentation). 
Note that the x-axis denotes the number of \emph{DI}s or CIs (in \%) used for performing Knowledge Distillation with respect to the training data size. 
}
\label{fig:ci-vs-di} 
\end{figure*}

\subsection{CIFAR-10}
\label{subsec:cifar}
Unlike MNIST and Fashion MNIST, this dataset contains RGB images of dimension $32 \times 32 \times 3$. The dataset contains $60000$ images from $10$ classes, where each class has $6000$ images. Among them, $50000$ images are form the training set and rest of the $10000$ images compose the test set. We take AlexNet~\cite{krizhevsky2012imagenet} as \Te{} model which is relatively large in comparison to LeNet-$5$. Since the standard AlexNet model is designed to process input of dimension $227 \times 227 \times 3$, we need to resize the input image to this large dimension. To avoid that, we have modified the standard AlexNet to accept $32 \times 32 \times 3$ input images. The modified AlexNet contains $5$ convolution layers with BatchNorm~\cite{batchnorm-icml-2015} regularization. Pooling is also applied on convolution layers $1, 2,$ and $5$. The deepest three layers are fully connected. AlexNet-Half is derived from the AlexNet by taking half of convolutional filters and half of the neurons in the fully connected layers except in the classification layer which has number of neurons equal to number of classes. The AlexNet-Half architecture is used as the \St{} model. The \Te{} and \St{} models have $1.65 \times 10^6$ and $7.23 \times 10^5$ parameters respectively. For architectural details of the teacher and the student nets, please refer to the supplementary document.

Table~\ref{tab:cifar} presents the results on the CIFAR-$10$ dataset. It can be observed that the proposed ZSKD approach can achieve knowledge distillation with the \emph{Data Impressions} that results in performance competitive to that realized using the actual data samples. Since the underlying target dataset is relatively more complex, we use a bigger transfer set containing $40000$ \emph{DI}s. However, the size of this transfer set containing \emph{DI}s is still $20\%$ smaller than that of the original training set size used for the classical knowledge distillation \cite{hinton2015distilling}.
\begin{table}[]
\caption{Performance of the proposed ZSKD framework on the CIFAR-$10$ dataset.}
\centering
\label{tab:cifar}
\begin{tabular}{|c|c|}
\hline
\multicolumn{1}{|c|}{\textbf{Model}} & \multicolumn{1}{c|}{\textbf{Performance}} \\ \hline
Teacher-CE                              &     83.03                                      \\ \hline
Student-CE                           &     80.04                                      \\ \hline
\makecell{Student-KD \cite{hinton2015distilling}\\ 50K original data}          &     80.08                                      \\ \hline
\makecell{\textbf{ZSKD} (Ours) \\($40000$ \emph{DI}s, and no original data)}                          &     69.56                                     \\ \hline
\end{tabular}
\end{table}

\subsection{Size of the Transfer Set}
\label{subsec:effect-size}
In this subsection, we investigate the effect of transfer set size on the performance of the distilled \St{} model. We perform the distillation with different number of \emph{Data Impressions} such as $\{1\%,\: 5\%,\: 10\%,\: \ldots, 80\%\}$ of the training set size. Figure~\ref{fig:data-vs-di} shows the performance of the resulting \St{} model on the test set for all the datasets. For comparison, the plots present performance of the models distilled with the equal number of actual training samples from the dataset. It is observed that, as one can expect, the performance increases with size of the transfer set. Interestingly, even a small number of \emph{Data Impressions} (e.g. $20\%$ of the training set size) are sufficient to provide a competitive performance, though the improvement in performance gets quickly saturated. Also, note that the initial performance (with smaller transfer set) reflects the complexity of the task (dataset). For simpler datasets such as MNIST, smaller transfer sets are sufficient to achieve competitive performance. In other words, small number of \emph{Data Impressions} can do the job of representing the patterns in the dataset. As the dataset becomes complex, more number of \emph{Data Impressions} need to be generated to capture the underlying patterns in the dataset. Note that similar trends are observed in the distillation with the actual training samples as well.

\subsection{Class Versus Data Impressions}
\label{subsec:class-data}
Feature visualization works such as~\cite{backprop-iclrw-2014,guidedbackprop-iclrw-2015,olah2017feature,deep-dream-2015} attempt to understand the patterns learned by the deep neural networks in order to recognize the objects. These works reconstruct a chosen neural activation in the input space as one way to explain away the model's inference.

One of the recent works by \cite{mopuri2018ask} reconstructs samples of a given class for a downstream task of adversarial fooling. They optimize a random noise in the input space till it results in a one-hot vector (softmax) output. This means, their optimization to craft the representative samples would expect a one-hot vector in the output space. Hence, they call the reconstructions \emph{Class Impressions}. Our reconstruction (eq.~(\ref{eqn:dir-ci})) is inspired from this, though we model the output space utilizing the class similarities perceived by the \Te{} model. Because of this, we argue that our modelling is closer to the original distribution and results in better patterns in the reconstructions, calling them \emph{Data Impressions} of the \Te{} model. 

In this subsection, we compare these two varieties of reconstructions for the application of distillation. Figure~\ref{fig:ci-vs-di} demonstrates the effectiveness of \emph{Class} and \emph{Data Impressions} over three datasets. It is observed that the proposed Dirichlet modelling of the output space and the reconstructed impressions consistently outperform their class counterparts by a large margin. Also, in case of \emph{Class Impressions}, the increment in the performance due to increased transfer set size is relatively small compared to that of \emph{Data Impressions}. Note that for better understanding, the results are shown without any data augmentation while conducting the distillation. 

\section{Discussion and Conclusion}
\label{sec:conclusion}
Knowledge Distillation~\cite{hinton2015distilling} and few-shot learning hold a great deal of potential in terms of both the challenges they pose and the applications that can be realised. Data-free learning presented in recent works such as \cite{dfkd-nips-lld-17,mopuri2018ask} can be treated as a type of zero-shot learning, where the aim is to extract or reconstruct samples of the underlying data distribution from a trained model in order to realize a target application. It is easy to see that this line of research has significant practical implications. For instance, a deep learned model can be obtained (i) from commercial products with deployed models (e.g., mobile phone or autonomous driving vehicle), or (ii) via hacking a deployment setup. In such cases only trained model is available without training data. Also, it can help us to mitigate the absence of training data in scenarios such as medical diagnosis, where, it is often the case that patients' privacy prohibits distribution of the training data. In those cases only the trained models can be made available. 
Further, given (i) the cost of annotating the data, and (ii) competitive advantage leveraged with more training data, it is quite a possibility that the trained models will be made available but not the actual training data. For example, models trained by Google and Facebook might utilize proprietary data such as JFT-300M, SFC.

In this work, we presented for the first time, a complete framework called Zero-Shot Knowledge Distillation (ZSKD) to perform knowledge distillation without utilizing any data samples or meta data extracted from it. We proposed a sample extraction mechanism via modelling the data distribution in the softmax space. As a useful prior, our model utilizes class similarity information extracted from the learned model and attempts to synthesize the underlying data samples. Further, we have investigated the effectiveness of the these synthesized samples, named \emph{Data Impressions} for a downstream task of training a substitute model via distillation.

A set of recent works that attempt to extract the training data from a learned model, drive a downstream task such as crafting adversarial perturbations or training a substitute model. However, in the current setup, the extracted samples are not influenced by the target task so as to call them task driven. Besides, it is not observed that these aforementioned extractions utilize any strong prior about the data distribution during the reconstruction. In that sense, our Dirichlet modelling of the output space that inculcates the visual similarity prior among the categories can be considered as a step towards a faithful extraction of the underlying patterns in the distribution. However, we believe that there is a lot of scope for imbibing additional and better priors particularly in the task driven scenario. For instance, utilizing multiple \Te{} models trained on different tasks can enable better extraction of the data patterns. Also, while estimating the impressions, we can formulate objectives that can explicitly encourage diversity in the extracted samples. Some of these ideas will be considered as our future research directions.

\bibliography{bibliography}
\bibliographystyle{conf}

\onecolumn
\begin{center}
    \LARGE{\textbf{Appendix}}
\end{center}

\setcounter{section}{0}
\section{Architecture Details Used in ZSKD}
\begin{itemize}
\item Lenet-$5$ as teacher and Lenet-$5$-Half as student used for MNIST and Fashion-MNIST datasets 
\end{itemize}

\begin{longtable}[c]{| m{15em} | m{15em}|}
\hline
\textbf{Lenet-5 Architecture} & \textbf{Lenet-5-Half Architecture} \\
(Teacher Model)  & (Student Model)\\
\hline
\textbf{Layer 1: Convolution } & \textbf{Layer 1: Convolution } \\
Input: 32x32x1; Output: 28x28x6  & Input: 32x32x1; Output: 28x28x3 \\
Kernel size: 5x5x1 (initialized through truncated normal with standard deviation of 0.1) 

No. Of Filters: 6, stride = 1 

Padding = ‘VALID’ 

Bias initialized with zeros. 

Activation: Relu & Kernel size: 5x5x1 (initialized through truncated normal with standard deviation of 0.1) 

No. Of Filters: 3, stride = 1 

Padding = ‘VALID’ 

Bias initialized with zeros. 

Activation: Relu \\
\hline
\textbf{Layer 2: Pooling } & \textbf{Layer 2: Pooling }\\
Max Pooling, Padding=’VALID’ 

Input: 28x28x6; Output: 14x14x6 & Max Pooling, Padding=’VALID’ 

Input: 28x28x3; Output: 14x14x3\\
\hline
\textbf{Layer 3: Convolution } & \textbf{Layer 3: Convolution }\\
Input: 14x14x6; Output: 10x10x16 

Kernel size: 5x5x6 (initialized through truncated normal with standard deviation of 0.1) 

No. Of Filters: 16, stride = 1 

Padding = ‘VALID’ 

Bias initialized with zeros. 

Activation: Relu & Input: 14x14x3; Output: 10x10x8 

Kernel size: 5x5x3 (initialized through truncated normal with standard deviation of 0.1) 

No. Of Filters: 8, stride = 1 

Padding = ‘VALID’ 

Bias initialized with zeros. 

Activation: Relu \\
\hline
\textbf{Layer 4: Pooling} & \textbf{Layer 4: Pooling} \\
Max Pooling, Padding=’VALID’ 

Input: 10x10x16; Output: 5x5x16 & Max Pooling, Padding=’VALID’ 

Input: 10x10x8; Output: 5x5x8 \\
\hline
\textbf{Flatten:} & \textbf{Flatten:} \\
Input: 5x5x16; Output=400 & Input: 5x5x8; Output=200 \\
\hline
\textbf{Layer 5: Fully Connected} & \textbf{Layer 5: Fully Connected} \\
Input: 400; Output:120 

Weight shape: (400,120) (initialized through truncated normal with standard deviation of 0.1) 

Bias initialized with zeros. 

Activation: Relu & Input: 200; Output:120 

Weight shape: (200,120) (initialized through truncated normal with standard deviation of 0.1) 

Bias initialized with zeros. 

Activation: Relu\\
\hline
\pagebreak
\hline
\textbf{Layer 6: Fully Connected} & \textbf{Layer 6: Fully Connected} \\
Input: 120; Output:84 

Weight shape: (120,84) (initialized through truncated normal with standard deviation of 0.1) 

Bias initialized with zeros. 

Activation: Relu & Input: 120; Output:84 

Weight shape: (120,84) (initialized through truncated normal with standard deviation of 0.1) 

Bias initialized with zeros. 

Activation: Relu \\
\hline
\textbf{Layer 7: Fully Connected} & \textbf{Layer 7: Fully Connected}\\

Input: 84; Output:10 

Weight shape: (84,10) (initialized through truncated normal with standard deviation of 0.1) 

Bias initialized with zeros. 

Output: Logits & Input: 84; Output:10 

Weight shape: (84,10) (initialized through truncated normal with standard deviation of 0.1) 

Bias initialized with zeros. 

Output: Logits \\
\hline
\textbf{Layer 8: Softmax Layer} & \textbf{Layer 8: Softmax Layer} \\
\hline

\caption{Teacher and Student Models for MNIST and Fashion-MNIST.\label{long}}\\

\end{longtable}

\begin{itemize}
\item Alexnet as Teacher and Alexnet-Half as student model used for CIFAR 10 dataset 
\end{itemize}
\begin{longtable}[c]{| m{15em} | m{15em}|}
\hline
\textbf{Alexnet Architecture} & \textbf{Alexnet-Half Architecture} \\
(Teacher Model)  & (Student Model)\\
\hline
\textbf{Layer 1: Convolution} & \textbf{Layer 1: Convolution} \\
Input: 32x32x3; Output: 32x32x48 

Kernel size: 5x5x3 (initialized through random normal with standard deviation of 0.01) 

No. Of Filters: 48, stride = 1 

Padding = ‘SAME’ 

Bias initialized with zeros. 

Activation: Relu & Input: 32x32x3; Output: 32x32x24 

Kernel size: 5x5x3 (initialized through random normal with standard deviation of 0.01) 

No. Of Filters: 24, stride = 1 

Padding = ‘SAME’ 

Bias initialized with zeros. 

Activation: Relu \\
\hline
\textbf{Layer 2: Local Response Normalization} & \textbf{Layer 2: Local Response Normalization} \\
depth\_radius =$2$, alpha =$0.0001$, beta=$0.75$, bias=$1.0$ &
depth\_radius =$2$, alpha =$0.0001$, beta=$0.75$, bias=$1.0$ \\
\hline
\textbf{Layer 3: Pooling} & \textbf{Layer 3: Pooling} \\
Max Pooling, Padding=’VALID’ 

Kernel size=3, stride =2  

Output: 15x15x48 & Max Pooling, Padding=’VALID’ 

Kernel size=3, stride =2  

Output: 15x15x24 \\
\hline
\textbf{Layer 4: Batch Norm} & \textbf{Layer 4: Batch Norm} \\
\hline
\textbf{Layer 5: Convolution} & \textbf{Layer 5: Convolution} \\
Input: 15x15x48; Output: 15x15x128 

Kernel size: 5x5x48 (initialized through random normal with standard deviation of 0.01) 

No. Of Filters: 128, stride = 1 

Padding = ‘SAME’ 

Bias initialized with 1.0 

Activation: Relu &  Input: 15x15x24; Output: 15x15x64 

Kernel size: 5x5x24 (initialized through random normal with standard deviation of 0.01) 

No. Of Filters: 64, stride = 1 

Padding = ‘SAME’ 

Bias initialized with 1.0 

Activation: Relu \\
\hline
\textbf{Layer 6: Local Response Normalization} & \textbf{Layer 6: Local Response Normalization} \\
depth\_radius =$2$, alpha =$0.0001$, beta=$0.75$, bias=$1.0$ &
depth\_radius =$2$, alpha =$0.0001$, beta=$0.75$, bias=$1.0$ \\
\hline
\textbf{Layer 7: Pooling} & \textbf{Layer 7: Pooling} \\
Max Pooling, Padding=’VALID’ 

Kernel size=3, stride =2  

Output: 7x7x128 & Max Pooling, Padding=’VALID’ 

Kernel size=3, stride =2  

Output: 7x7x64 \\
\hline
\textbf{Layer 8: Batch Norm} & \textbf{Layer 8: Batch Norm} \\
\hline
\textbf{Layer 9: Convolution} & \textbf{Layer 9: Convolution} \\
Input: 7x7x128; Output: 7x7x192 

Kernel size: 3x3x128 (initialized through random normal with standard deviation of 0.01) 

No. Of Filters: 192, stride = 1 

Padding = ‘SAME’ 

Bias initialized with zeros. 

Activation: Relu & Input: 7x7x64; Output: 7x7x96 

Kernel size: 3x3x64 (initialized through random normal with standard deviation of 0.01) 

No. Of Filters: 96, stride = 1 

Padding = ‘SAME’ 

Bias initialized with zeros. 

Activation: Relu \\
\hline
\textbf{Layer 10: Batch Norm} & \textbf{Layer 10: Batch Norm} \\
\hline
\textbf{Layer 11: Convolution} & \textbf{Layer 11: Convolution} \\
Input: 7x7x192; Output: 7x7x192 

Kernel size: 3x3x192 (initialized through random normal with standard deviation of 0.01) 

No. Of Filters: 192, stride = 1 

Padding = ‘SAME’ 

Bias initialized with 1.0. 

Activation: Relu & Input: 7x7x96; Output: 7x7x96 

Kernel size: 3x3x96 (initialized through random normal with standard deviation of 0.01) 

No. Of Filters: 96, stride = 1 

Padding = ‘SAME’ 

Bias initialized with 1.0. 

Activation: Relu \\
\hline
\textbf{Layer 12: Batch Norm} & \textbf{Layer 12: Batch Norm} \\
\hline
\textbf{Layer 13: Convolution} & \textbf{Layer 13: Convolution} \\
Input: 7x7x192; Output: 7x7x128 

Kernel size: 3x3x192 (initialized through random normal with standard deviation of 0.01) 

No. Of Filters: 128, stride = 1 

Padding = ‘SAME’ 

Bias initialized with 1.0. 

Activation: Relu & Input: 7x7x96; Output: 7x7x64 

Kernel size: 3x3x96 (initialized through random normal with standard deviation of 0.01) 

No. Of Filters: 64, stride = 1 

Padding = ‘SAME’ 

Bias initialized with 1.0. 

Activation: Relu \\
\hline
\textbf{Layer 14: Pooling} & \textbf{Layer 14: Pooling} \\
Max Pooling, Padding=’VALID’ 

Kernel size=3, stride =2  

Output: 3x3x128 & Max Pooling, Padding=’VALID’ 

Kernel size=3, stride =2  

Output: 3x3x64 \\
\hline
\textbf{Layer 15: Batch Norm} & \textbf{Layer 15: Batch Norm} \\
\hline
\textbf{Flatten: } & \textbf{Flatten: } \\
Input: 3x3x128; Output=1152 & Input: 3x3x64; Output=576 \\
\hline
\textbf{Layer 16: Fully Connected } & \textbf{Layer 16: Fully Connected } \\
Input: 1152; Output:512 

Weight shape: (1152,512) (initialized through random normal with standard deviation of 0.01) 

Bias initialized with zeros. 

Activation: Relu & Input: 576; Output:256 

Weight shape: (576,256) (initialized through random normal with standard deviation of 0.01) 

Bias initialized with zeros. 

Activation: Relu \\
\hline
\textbf{Layer 17: Dropout} & \textbf{Layer 17: Dropout}\\
Rate=0.5 & Rate=0.5 \\
\hline
\textbf{Layer 18: Batch Norm} & \textbf{Layer 18: Batch Norm} \\
\hline
\pagebreak
\hline
\textbf{Layer 19: Fully Connected} & \textbf{Layer 19: Fully Connected} \\
Input: 512; Output:256 

Weight shape: (512,256) (initialized through random normal with standard deviation of 0.01) 

Bias initialized with zeros. 

Activation: Relu & Input: 256; Output:128 

Weight shape: (256,128) (initialized through random normal with standard deviation of 0.01) 

Bias initialized with zeros. 

Activation: Relu \\
\hline
\textbf{Layer 20: Dropout} & \textbf{Layer 20: Dropout} \\
Rate=0.5 & Rate=0.5 \\
\hline
\textbf{Layer 21: Batch Norm} & \textbf{Layer 21: Batch Norm} \\
\hline
\textbf{Layer 22: Fully Connected} &  \textbf{Layer 22: Fully Connected} \\
Input: 256; Output:10 

Weight shape: (256,10) (initialized through random normal with standard deviation of 0.01) 

Bias initialized with zeros. 

Output: Logits  & Input: 128; Output:10 

Weight shape: (128,10) (initialized through random normal with standard deviation of 0.01) 

Bias initialized with zeros. 

Output: Logits \\
\hline
\textbf{Layer 23: Softmax Layer} & \textbf{Layer 23: Softmax Layer} \\
\hline
\caption{Teacher and Student Models for CIFAR 10.\label{long}}\\
\end{longtable}

\textit{Note: During Distillation, at train time Logits are divided by temperature of 20 and at test time the Logits are divided by temperature of 1. }

\section{Details of Hyperparameters Used in ZSKD}
NOTE:- All the experiments are performed using TensorFlow framework.
\subsection{MNIST Training}
Teacher Model: Lenet-5 \\
Student Model: Lenet-5-Half

\begin{description}
  \item[$\bullet$ Teacher Training with original data: ] We take epochs as 200, batch size of 512, learning rate equal to 0.001 and Adam optimizer. 
  \item[$\bullet$ Student Training with original data using cross entropy loss:] Same hyperparameters as above. 
  \item[$\bullet$ Student Training with original data using knowledge distillation: ] We take $\lambda$ = 0.3 which is the weight given to cross entropy loss and the distillation loss is given the weight as 1.0. The learning rate is taken as 0.01, temperature as 20 and rest of the hyperparameters are same. 
  
  \item[$\bullet$ Data Impressions (DI) Generation:] 
 
\end{description}
\begin{enumerate}[label=(\alph*)]
\item \textbf{1\% (600 DI): } Batch size of 10, number of iterations to be 1500 and learning rate as 0.1 
\item \textbf{5\% (3000 DI): }  Batch size as 10, number of iterations to be 1500 and learning rate as 0.1 
\item \textbf{10\% (6000 DI): }  Batch size as 100, number of iterations to be 1500 and learning rate as 1.0 
\item \textbf{20\% (12000 DI): } Batch size as 100, number of iterations to be 1500 and learning rate as 2.0 
\item \textbf{40\% (24000 DI): } Batch size as 100, number of iterations to be 1500 and learning rate as 3.0 
\end{enumerate}
\textbf{Student Training using DI (end to end):}We take learning rate of 0.01, batch size as 512, max epochs to be 2000 and Adam optimizer. \\ \\
We further finetune the model pretrained on 40\% DI using mixture of DI and augmented DI samples with learning rate of 0.001 
\begin{description}
    \item[$\bullet$  Class Impressions (CI) Generation:]
\end{description}

NOTE: We randomly sample a value (say x) from confidence range of 0.55 and 0.70. The training is done on the random noisy image till the confidence of noisy image $\textgreater$ = confidence of x  

\begin{enumerate}[label=(\alph*)]
\item \textbf{1\% (600 DI): } We take learning rate as 2.0 and student trained with learning rate of 0.01 
\item \textbf{5\% (3000 DI): }  We take learning rate as 0.01 and student trained with learning rate of 0.01 
\item \textbf{10\% (6000 DI): }  We take learning rate as 0.1 and student trained with learning rate of 0.01 
\item \textbf{20\% (12000 DI): } We take learning rate as 0.01 and student trained with learning rate of 0.01  
\item \textbf{40\% (24000 DI): } We take learning rate as 0.1 and student trained with learning rate of 0.001 
\end{enumerate}

\subsection{ Fashion - MNIST Training}
Teacher Model: Lenet-5 \\
Student Model: Lenet-5-Half

\begin{description}
  \item[$\bullet$ Teacher Training with original data: ] We take epochs as 200, batch size of 512, learning rate equal to 0.001 and Adam optimizer. 
  \item[$\bullet$ Student Training with original data using cross entropy loss:] Same hyperparameters as above. 
  \item[$\bullet$ Student Training with original data using knowledge distillation: ] We take $\lambda$ = 0.3 which is the weight given to cross entropy loss and the distillation loss is given the weight as 1.0. The learning rate is taken as 0.01, temperature as 20 and rest of the hyperparameters are same. 
  
  \item[$\bullet$ Data Impressions (DI) Generation:] 
 
\end{description}
\begin{enumerate}[label=(\alph*)]
\item \textbf{1\% (600 DI): } Batch size of 10, number of iterations to be 1500 and learning rate as 3.0
\item \textbf{5\% (3000 DI): }  Batch size as 10, number of iterations to be 1500 and learning rate as 3.0
\item \textbf{10\% (6000 DI): }  Batch size as 100, number of iterations to be 1500 and learning rate as 1.0
\item \textbf{20\% (12000 DI): } Batch size as 100, number of iterations to be 1500 and learning rate as 1.0
\item \textbf{40\% (24000 DI): } Batch size as 10, number of iterations to be 1500 and learning rate as 1.0 
\item \textbf{80\% (48000 DI): } Batch size as 100, number of iterations to be 1500 and learning rate as 3.0 
\end{enumerate}
\textbf{Student Training using DI (end to end):} We take batch size as 512, max epochs to be 2000 and Adam optimizer. Learning rate are taken as follows:

    \begin{description}
        \item[$\bullet$]Learning rate as 0.01 in case of (a), (d), (e) and (f).
        \item[$\bullet$]Learning rate as 0.001 in case of b).
        \item[$\bullet$]Learning rate as 0.0001 in case of c).
    \end{description}
We further finetune the model pretrained on 80\% DI using mixture of DI and augmented DI samples with learning rate of 0.001 
\\
\begin{description}
    \item[$\bullet$  Class Impressions (CI) Generation:]
\end{description}

NOTE: We randomly sample a value (say x) from confidence range of 0.55 and 0.70. The training is done on the random noisy image till the confidence of noisy image $\textgreater$ = confidence of x  
\\
\begin{enumerate}[label=(\alph*)]
\item \textbf{1\% (600 DI): } We take learning rate as 0.01 and student trained with learning rate of 0.001 
\item \textbf{5\% (3000 DI): }  We take learning rate as 0.1 and student trained with learning rate of 0.001 
\item \textbf{10\% (6000 DI): } We take learning rate as 2.0 and student trained with learning rate of 0.001 
\item \textbf{20\% (12000 DI): } We take learning rate as 1.0 and student trained with learning rate of 0.001 
\item \textbf{40\% (24000 DI): } We take learning rate as 0.01 and student trained with learning rate of 0.01 
\item \textbf{80\% (48000 DI): } We take learning rate as 0.5 and student trained with learning rate of 0.001 
\end{enumerate}

\subsection{ CIFAR 10 Training}
Teacher Model: Alexnet \\
Student Model: Alexnet-Half

\begin{description}
  \item[$\bullet$ Teacher Training with original data: ] We take epochs as 1000, batch size of 512, learning rate equal to 0.001 and Adam optimizer. 
  \item[$\bullet$ Student Training with original data using cross entropy loss:] Same hyperparameters as above. 
  \item[$\bullet$ Student Training with original data using knowledge distillation: ] We take $\lambda$ = 0.3 which is the weight given to cross entropy loss and the distillation loss is given the weight as 1.0. The learning rate is taken as 0.001, temperature as 20 and rest of the hyperparameters are same. 
  
  \item[$\bullet$ Data Impressions (DI) Generation:] 
 
\end{description}
\begin{enumerate}[label=(\alph*)]
\item \textbf{1\% (500 DI): } Batch size of 5, number of iterations to be 1500 and learning rate as 0.01 
\item \textbf{5\% (2500 DI): }  Batch size as 25, number of iterations to be 1500 and learning rate as 0.01 
\item \textbf{10\% (5000 DI): }   Batch size as 50, number of iterations to be 1500 and learning rate as 0.01 
\item \textbf{20\% (10000 DI): }Batch size as 100, number of iterations to be 1500 and learning rate as 0.01 
\item \textbf{40\% (20000 DI): }  Batch size as 100, number of iterations to be 1500 and learning rate as 0.01 
\item \textbf{80\% (40000 DI): } Batch size as 100, number of iterations to be 1500 and learning rate as 0.01 
\end{enumerate}
\textbf{Student Training using DI (end to end):} We take learning rate of 0.001, batch size as 512, max epochs to be 2000 and Adam optimizer. 

We further finetune the model pretrained on 80\% DI using mixture of DI and augmented DI samples with learning rate of 0.001 having batch size as 5000.
\pagebreak
\begin{description}
    \item[$\bullet$  Class Impressions (CI) Generation:]
\end{description}

NOTE: We randomly sample a value (say x) from confidence range of 0.55 and 0.70. The training is done on the random noisy image till the confidence of noisy image $\textgreater$ = confidence of x  
\\
\begin{enumerate}[label=(\alph*)]
\item \textbf{1\% (500 DI): } We take learning rate as 0.1 and student trained with learning rate of 0.001 
\item \textbf{5\% (2500 DI): }  We take learning rate as 0.1 and student trained with learning rate of 0.01 
\item \textbf{10\% (5000 DI): }We take learning rate as 0.1 and student trained with learning rate of 0.01 
\item \textbf{20\% (10000 DI): } We take learning rate as 2.0 and student trained with learning rate of 0.001 
\item \textbf{40\% (20000 DI): } We take learning rate as 1.0 and student trained with learning rate of 0.001 
\item \textbf{80\% (40000 DI): } We take learning rate as 0.1 and student trained with learning rate of 0.01 
\end{enumerate}
\section{Details on Augmentation}
The following operations are done on the DI's to create variety of augmented samples :- 

\begin{enumerate}[label=(\roman*)]
\item Scaling of 90\%, 75\% and 60\% of original DI's
\item Translation is done on left, right, top and bottom directions by 20\%
\item Rotation: Starts at -90$^{\circ}$ and ends at +90$^{\circ}$ to produce 10 rotated DI's such that the degree of next rotation is 20$^{\circ}$ more than the previous angle of rotation
\item Flipping: Operations done are flip left right, flip up down and transpose
\item Scaling and Translation: The scaled Di's are translated on left, right, top and bottom directions by 20\%
\item Translation and Rotation: The translated Di's are rotated 
\item Scaling and Rotation: The scaled DI's are rotated
\end{enumerate}
Below three operations are further exclusively done on the DI's extracted from Alexnet teacher model. These DI's have RGB components whereas the DI's obtained from Lenet teacher are gray scaled.
\begin{itemize}
\item Salt and Pepper Noise
\item Gaussian Noise
\item Adding Gaussian Noise to Salt and Pepper Noised DI
\end{itemize}

\begin{center}
    \large{\textbf{Ablations: With and without Augmentation}}
\end{center}
\begin{table}[h]

\label{tab:aug_ablations_on_student}
\centering
\begin{tabular}{|c|c|c|}
\hline
Teacher Model trained on Data set    & \multicolumn{2}{c|}{ZSKD Performance on Student Network}\\ \cline{2-3}
 &  {Without Augmentation} & {With Augmentation} \\
\hline
\hline
MNIST                              &     96.98              &       \bf{98.77}                 \\ \hline
Fashion MNIST                           &     69.37          &      \bf{79.62}                     \\ \hline
CIFAR 10                           &     56.80               &      \bf{69.56}                \\ \hline
\end{tabular}
\caption{Performance (in \%) of the proposed ZSKD framework.}
\end{table}
\pagebreak
\begin{center}
    \large{\textbf{Uniform Prior v/s Class Similarity Prior}}
\end{center}
\begin{table}[h]

\label{tab:aug_ablations_on_student}
\centering
\begin{tabular}{|c|c|c|}

\hline
Dataset   & Uniform Prior & Class Similarity Prior \\
\hline
\hline

MNIST                              &     95.16              &       96.98                 \\ \hline
Fashion MNIST                           &     56.24          &      69.37                     \\ \hline
Cifar 10                           &     49.23               &      56.80               \\ \hline
\end{tabular}
\caption{Performance of proposed ZSKD (in \%) using uniform and class similarity priors (without augmentation)}
\end{table}

\end{document}